\documentclass[runningheads,a4paper]{llncs}

\usepackage{amssymb}
\usepackage{graphicx}
\usepackage{amsmath}
\usepackage{booktabs}
\usepackage{color}
\usepackage{todonotes}
\usepackage[utf8x]{inputenc}

\usepackage{url}
\newcommand{\keywords}[1]{\par\addvspace\baselineskip
\noindent\keywordname\enspace\ignorespaces#1}

\begin{document}

\title{RDF2Vec-based Classification of Ontology Alignment Changes}
\author{Matthias Jurisch, Bodo Igler}
\institute{
RheinMain University of Applied Sciences\\
Department of Design -- Computer Science -- Media\\
Unter den Eichen 5\\
65195 Wiesbaden, Germany\\
\email{matthias.jurisch@hs-rm.de, bodo.igler@hs-rm.de}
}
\maketitle

\begin{abstract}

When ontologies cover overlapping topics, the overlap can be represented using 
	ontology alignments. These alignments need to be continuously adapted 
	to changing ontologies. Especially for large ontologies this is a 
	costly task often consisting of manual work.  Finding changes that do 
	not lead to an adaption of the alignment can potentially make this 
	process significantly easier. This work presents an approach to finding 
	these changes based on RDF embeddings and common classification 
	techniques.  To examine the feasibility of this approach, an evaluation 
	on a real-world dataset is presented. In this evaluation, the best 
	classifiers reached a precision of 0.8.
\end{abstract}


\section{Introduction}
\label{sec:Introduction}

Finding alignments between ontologies, also known as ontology matching, is a 
non-trivial task and has been an active area of research over the last ten 
years. Several approaches in this area are based on the structure of the 
ontologies, logical axioms or lexical similarity \cite{Euzenat2007}.  However, 
once these alignments are found, they will not necessarily stay untouched 
forever. Especially when alignments connect large ontologies, adapting these 
alignments to changes is a work-intensive task. In the area of biomedical 
ontologies, some alignments contain around 6500 correspondences that might be 
affected by a change in one of the ontologies they connect. Given a change in 
the ontology, detecting which parts of the alignment are affected by the change 
and need to be adapted is not a trivial task that usually requires manual 
labour. The effort required for this task can be significantly reduced, if some 
changes can be excluded from it.  However, it is usually not clear how to 
identify changes that do not
affect the alignment.

In this paper, we propose an approach to this problem based on RDF embeddings 
and well-known classification techniques. The central aspect of this approach 
is to represent changed concepts by their RDF embedding and classify whether an 
alignment statement nearby should be changed. To gain evidence if this approach 
works, we evaluate it using a dataset from the area of biomedical ontologies.  
On this dataset, our approach is able to identify changes affecting alignment 
statements with a precision of 0.8.

The remainder of this work is structured as follows: Section \ref{sec:Related 
Work} discusses foundations of our work and related approaches. The general 
approach is presented in Section \ref{sec:Approach}. Evaluation methodology, 
the dataset and results are shown in Section \ref{sec:Evaluation}. Section 
\ref{sec:Discussion} discusses the results of our evaluation, and advantages 
and disadvantages of our approach. A conclusion is given in Section 
\ref{sec:Conclusion}.

\section{Foundations and Related Work} 
\label{sec:Related Work}

An ontology alignment (sometimes also called ontology mapping) is a set of 
correspondences between entities in different ontologies \cite{Euzenat2007}.
To make it easier to reason about these alignments, we use the following formal 
definition in the style of \cite{Gross2013} for ontology mappings: An alignment 
between two ontologies $\mathcal{O}_1$ and $\mathcal{O}_2$ is defined as
\[A_{\mathcal{O}_1,\mathcal{O}_2}=\{(c_1, c_2, semType)| c_1 \in \mathcal{O}_1, 
c_2 \in \mathcal{O}_2, semType \in \{\equiv, \leq, \geq\}\}\]

$A_{\mathcal{O}_1,\mathcal{O}_2}$ is the set of all \emph{alignment 
statements}.  To denote a change of an ontology over time, we use the prime 
symbol (e.g., a changed version of $\mathcal{O}$ is denoted as $\mathcal{O}'$). 
The alignment adaption problem for two ontologies $\mathcal{O}_1$ and 
$\mathcal{O}_2$ connected by $A_{\mathcal{O}_1,\mathcal{O}_2}$ can then be 
stated as finding a new alignment $A'_{\mathcal{O}'_1,\mathcal{O}'_2}$, when  
$\mathcal{O}_1$ and
$\mathcal{O}_2$ evolve to $\mathcal{O}'_1$ and $\mathcal{O}'_2$.

In the area of ontology alignment adaption, several approaches are based on 
rules or rule-based dependency analysis. \cite{Klein2003} is focussed on 
finding which changes are relevant to parts of the alignment using a dependency 
analysis. \cite{Velegrakis2003} proposed an \emph{incremental approach}  
reacting to specific changes in database schemas based on rules. For each 
change pattern a specific modification for the mapping is defined.  
\cite{Yu2005} proposed an approach that is based on a composition of 
alignments.  A new alignment $A'_{\mathcal{O}'_1,\mathcal{O}'_2}$ is created by 
a \emph{composition of the alignment} $A_{\mathcal{O}_1,\mathcal{O}_2}$ and  
$A^+_{\mathcal{O}_2,\mathcal{O}'_2}$, the
alignment between $\mathcal{O}_2$ and $\mathcal{O}'_2$.
\cite{Gross2013} have shown that these techniques can also be applied to 
ontologies. However, all of these approaches require a set of rules that need 
to be constructed by a domain expert and are not necessarily reusable for 
other domains. Also, these approaches are not able to identify which changes in 
the ontologies are prone to causing an alignment change.

The task of knowledge base completion shares some properties with the problem 
we address in this work. In that area, classifiers are given a subject and a 
predicate and try to predict an object \cite{Nickel2016}. Approaches like 
\cite{Yang2014}, \cite{Joulin2017} and \cite{Socher2013} also use vector 
representations for prediction. However, this task does not take changes in the 
knowledge bases into account and is not applied to ontology alignments.

To our knowledge, no approach exists that predicts whether a given change has 
an impact on the alignment without using a detailed set of rules. This issue is 
at the core of our research.

\section{Approach}
\label{sec:Approach}

Our general approach is based on the representation of changed resources using 
RDF embeddings, a represenation of RDF nodes as vectors in a high-dimensional, 
dense vector space. RDF embeddings are generated using RDF2Vec \cite{RDF2Vec}, 
an approach  based on random graph walks as input to Word2Vec 
\cite{Mikolov2013}. The RDF2Vec-Model is trained on an RDF graph consisting of 
the ontologies $\mathcal{O}_1$ and $\mathcal{O}_2$ as well as the alignment 
$A_{\mathcal{O}_1 ,\mathcal{O}_2}$ as defined in section \ref{sec:Related 
Work}. With these embeddings, we train a classifier on whether a changed 
resource affected an alignment statement and use this classifier to predict 
whether other changes will affect the alignment.  We define a changed resource 
to lead to an alignment change, if a changed alignment statement is within a 
distance of two in the RDF graph.  This relatively small measure is used to 
make it easier to exclude certain regions from the search for affected 
statements.  For the same reason, only changes that are close to an alignment 
axiom are regarded.  The respective changes $c$ are extracted using an 
extension of the Protégé plugin 
owl-diff\footnote{\url{https://github.com/mhfj/owl-diff}}.  By comparing the 
parts of $A_{\mathcal{O}_1 ,\mathcal{O}_2}$ and $A'_{\mathcal{O}'_1 
,\mathcal{O}'_2}$ that are in the direct neighbourhood of $c$, it is possible 
to separate all changes into two groups: (1) changes that caused an alignment 
change in their neighbourhood and (2) changes that did not cause an alignment 
change in their neighbourhood and therefore did not affect the alignment.  

Each changed resource is represented by the corresponding RDF2Vec vector.  
Hence, the input to the training of the classifier is a pair $(v(c), k)$ 
consisting of a vector $v(c)$ and a class $k$. $k$ determines whether $c$ 
caused a change in its direct neighbourhood. The task at hand is to correctly 
classify new vectors.
To solve this problem, we use several common classification techniques: 
Regression, Naive Bayes, Tree-Based Algorithms as well as Support Vector 
Machines and Multilayer Perceptrons. Each algorithm is trained on one set of 
changes and evaluated on a different set.

\section{Evaluation}
\label{sec:Evaluation}

The research questions behind our evaluation are the following:

\begin{enumerate}
	\item Can RDF embeddings be used for change classification with an 
		acceptable performance? This question tries to clarify, whether 
		our approach is in general applicable to the problem at hand.
	\item Which classifiers can be used for this problem? This question is 
		used to identify the best classifiers for our problem.
\end{enumerate}

\subsection{Dataset}
\label{sec:Dataset}

The dataset used to answer these research questions  in our experiments is a 
real-word dataset from the domain of biomedical ontologies.  It has been used 
in several works that deal with alignment adaption, e.g., 
\cite{JimenezRuiz2010}, \cite{Gross2013}. The dataset comprises three 
ontologies: SNOMED-CT, the NCI-Thesaurus and FMA. For each ontology, yearly 
versions from 2009-2012 are available. Additionally, the dataset contains 
alignments extracted from the UMLS metathesaurus between the ontologies for 
each year.  This dataset has been made publicly 
available\footnote{\url{https://dbs.uni-leipzig.de/de/research/projects/evolution_of_ontologies_and_mappings/ontology_mapping_adaption}} 
by the authors of \cite{Gross2013}.

For simplicity of our presentation, we will only present the alignment between 
the ontologies NCI and FMA in the version change from 2009 to 2010. In the 
formal notation introduced in Section \ref{sec:Related Work}, $\mathcal{O}_1$ 
and $\mathcal{O}_2$ refer to the ontologies NCI and FMA as of 2009 and 
$\mathcal{O}'_1$ and $\mathcal{O}'_2$ as of 2010, respectively.  The alignment 
from 2009 is denoted by $A_{\mathcal{O}_1,\mathcal{O}_2}$ and the version from 
2010 by $A'_{\mathcal{O}'_1,\mathcal{O}'_2}$.

From 2009 to 2010, 924 changes are near alignment statements of which 47\% 
require an adaption.  These changes are used as a training set. The test set 
consists of the changes from 2010 to 2011.  This set contains 785 changes near 
alignment statements, of which 36\% lead to an alignment adaption.

\subsection{Methodology}

To generate RDF embeddings, the code from RDF2Vec \cite{RDF2Vec} was used.  The 
embeddings were trained using the skip gram model, with 500 dimensions used for 
the embeddings and random walks of length 8, as this was identified as the 
best-performing variant in \cite{RDF2Vec}.  An overview regarding 
classification methods used on these embeddings is given in Table 
\ref{tab:Classifiers}. Standard 
scikit\footnote{\url{http://scikit-learn.org/stable/}} implementations are used 
for the classification process. The classifiers are trained on changes from 
2009-2010 and validated on changes from 2010-2011 of the dataset described in 
Section \ref{sec:Dataset}. The performance of different classification 
techniques is evaluated based on \emph{f1-measure}, \emph{accuracy}, 
\emph{precision} and \emph{recall}.

\begin{table}[htb]
\caption{Classifiers}
\centering
	\label{tab:Classifiers}
	\begin{tabular}{ll}
	\toprule
		Category & Method\\
	\midrule
		Regression & Logistic Regression (LR)\\
		Naive Bayes & Gaussian Naive Bayes (NB)\\
		Nearest Neighbour & KNN\\
		Tree-Based Algorithms & CART,  Random Forest\\
		Support Vector Machines & RBF-Kernel, Linear Kernel\\
		Multilayer Perceptron & MLP hiden-layer-size: 250; 250,250;
		500; 500,500\\
	\bottomrule
\end{tabular}
\end{table}

\subsection{Results}
\label{sec:Results}

The results of the described process are displayed in Table \ref {tab:Results}.  
Only changes close to the alignment were included in this evaluation, since it 
would otherwise be very easy to achieve accuracy values above 95\%. Results for 
MLP did not vary based on the structure of the hidden layers, so one row 
represents all MLP results.  All algorithms show a very similar performance 
regarding the evaluated metrics.  The highest achieved precision is 0.81, which 
can be reached using MLP and linear SVM classification. These methods also 
reach the highest f1-measures of 0.75.  Accuracy of all classifiers is only 
marginally higher than what can be achieved using random guessing, given the 
distribution of classes in the test set. 

\begin{table}[htb]
\caption{Classification Results}
\centering
	\label{tab:Results}
	\begin{tabular}{l@{\qquad} c@{\qquad} c@{\qquad} c@{\qquad} c@{\qquad} 
		c}
	\toprule
		 & \emph{f1-measure} & \emph{accuracy} & \emph{precision} 
		& \emph{recall} \\
	\midrule
LR & 0.74 & 0.67 & 0.80 & 0.69  \\
NB & 0.67 & 0.58 & 0.73 & 0.62  \\
KNN & 0.71 & 0.62 & 0.75 & 0.69  \\
CART & 0.73 & 0.65 & 0.80 & 0.68  \\
RandomForest & 0.75 & 0.67 & 0.80 & 0.70  \\
SVM rbf & 0.74 & 0.65 & 0.77 & 0.71  \\
SVM linear & 0.75 & 0.67 & 0.81 & 0.70  \\
MLP & 0.75 & 0.68 & 0.81 & 0.70  \\ 

	\bottomrule
\end{tabular}
\end{table}

\section{Discussion}
\label{sec:Discussion}

The results presented in section \ref{sec:Results} give us some evidence on our 
first research question: Using RDF embeddings to represent changes seems to be 
a promising approach to the mapping adaption problem, as we can see a precision 
around 0.8. In general, several classification approaches show a similar 
performance. This precision can be achieved, although the approach uses no 
information regarding the nature of changes, e.g., the algorithm can not 
distinguish the correction of typos from major, structural changes.

An important advantage of this approach is that no sophisticated change model 
that is adapted to the domain is required. Approaches like \cite{Gross2013} 
require a rule-base that needs to be constructed from a detailed understanding 
of typical changes in the domain the ontologies describe. Hence, the author of 
these rules needs to be an expert in ontology engineering as well as the 
application domain. Also, these rules need to be constantly adapted to evolving 
domains, whereas an RDF-embedding based approach could learn new patterns 
autonomously.
However, to demonstrate these advantages, it is still required to show that 
this approach is also applicable to other data sets and different application 
domains.

\section{Conclusion and Outlook}
\label{sec:Conclusion}

In this work, we presented an approach to ontology alignment adaption based on 
RDF embeddings and common classification techniques. An evaluation on a dataset 
from the biomedical domain provided some evidence, that the approach is 
feasible. On the dataset, best-performing classifiers had a precision of 0.8. 

As future work, several extensions are possible: Further evaluations could be 
performed on different datasets. Also, a combination of this approach with 
existing mapping adaption approaches could be examined.  Change types could be 
used as another input to the classification process to improve classification 
accuracy. Another aspect for future work is to determine, when embeddings need 
to be updated, since embeddingis will become outdated when ontologies change.

\bibliographystyle{plain}
\bibliography{paperDL4KGS}

\begin{thebibliography}{10}

\bibitem{Euzenat2007}
J\'{e}r\^{o}me Euzenat and Pavel Shvaiko.
\newblock {\em {Ontology matching}}.
\newblock Springer, Heidelberg, 2007.

\bibitem{Gross2013}
Anika Gro{\ss}, Julio~Cesar {dos Reis}, Michael Hartung, C{\'{e}}dric Pruski,
  and Erhard Rahm.
\newblock {Semi-automatic adaptation of mappings between life science
  ontologies}.
\newblock {\em Lecture Notes in Computer Science (including subseries Lecture
  Notes in Artificial Intelligence and Lecture Notes in Bioinformatics)}, 7970
  LNBI:90--104, 2013.

\bibitem{JimenezRuiz2010}
Ernesto Jiménez-ruiz, Bernardo~Cuenca Grau, Ian Horrocks, and Rafael Berlanga.
\newblock Logic-based assessment of the compatibility of {{UMLS}} ontology
  sources.
\newblock In {\em JOURNAL OF BIOMEDICAL SEMANTICS}, 2010.

\bibitem{Joulin2017}
Armand Joulin, Edouard Grave, Piotr Bojanowski, Maximilian Nickel, and Tomas
  Mikolov.
\newblock {Fast Linear Model for Knowledge Graph Embeddings}.
\newblock 2017.

\bibitem{Klein2003}
Michel Klein and Heiner Stuckenschmidt.
\newblock {Evolution Management for Interconnected Ontologies}.
\newblock {\em Workshop on Semantic Integration at ISWC 2003}, 2003.

\bibitem{Mikolov2013}
Tomas Mikolov, Kai Chen, Greg Corrado, and Jeffrey Dean.
\newblock Efficient estimation of word representations in vector space.
\newblock {\em CoRR}, abs/1301.3781, 2013.

\bibitem{Nickel2016}
Maximilian Nickel, Kevin Murphy, Volker Tresp, and Evgeniy Gabrilovich.
\newblock A review of relational machine learning for knowledge graphs.
\newblock {\em Proceedings of the IEEE}, 104:11--33, 2016.

\bibitem{RDF2Vec}
Petar Ristoski and Heiko Paulheim.
\newblock {RDF2Vec}: {RDF} graph embeddings for data mining.
\newblock In {\em The Semantic Web - {ISWC} 20162016}, pages 498--514, 2016.

\bibitem{Socher2013}
Richard Socher, Danqi Chen, Christopher Manning, Danqi Chen, and Andrew Ng.
\newblock {Reasoning With Neural Tensor Networks for Knowledge Base
  Completion}.
\newblock {\em Neural Information Processing Systems (2003)}, pages 926--934,
  2013.

\bibitem{Velegrakis2003}
Yannis Velegrakis, Ren{\'{e}}e~J. Miller, and Lucian Popa.
\newblock {Mapping adaptation under evolving schemas}.
\newblock {\em VLDB '03 Proceedings of the 29th international conference on
  Very large data bases - Volume 29}, pages 584--595, 2003.

\bibitem{Yang2014}
Bishan Yang, Wen-tau Yih, Xiaodong He, Jianfeng Gao, and Li~Deng.
\newblock {Embedding Entities and Relations for Learning and Inference in
  Knowledge Bases}.
\newblock 2014.

\bibitem{Yu2005}
Cong Yu and Lucian Popa.
\newblock {Semantic Adaptation of Schema Mappings when Schemas Evolve}.
\newblock {\em Very Large Data Bases}, pages 1006 -- 1017, 2005.

\end{thebibliography}
\end{document}